\documentclass[conference]{IEEEtran}
\IEEEoverridecommandlockouts
\ifCLASSOPTIONcompsoc
\usepackage[nocompress]{cite}
\else
\usepackage{cite}
\fi
\usepackage{amsmath,amssymb,amsfonts}
\usepackage{algorithmic}
\usepackage{graphicx}
\usepackage{textcomp}
\usepackage{xcolor}
\usepackage{multirow}
\usepackage[T5]{fontenc}

\def\BibTeX{{\rm B\kern-.05em{\sc i\kern-.025em b}\kern-.08em
    T\kern-.1667em\lower.7ex\hbox{E}\kern-.125emX}}
\begin{document}

\title{AdaCS: Adaptive Normalization for Enhanced Code-Switching ASR\\
}



\author{\IEEEauthorblockN{The Chuong Chu\IEEEauthorrefmark{1}\IEEEauthorrefmark{6}, Vu Tuan Dat Pham\IEEEauthorrefmark{2}\IEEEauthorrefmark{6}, Trung Kien Dao\IEEEauthorrefmark{3}, Ngoc Hoang Nguyen\IEEEauthorrefmark{4} and Steven Truong\IEEEauthorrefmark{5}}
\IEEEauthorblockA{Department of Applied Scientist\\
VinBrain, Hanoi, Vietnam\\ 
\{\IEEEauthorrefmark{1}chuong.chu, \IEEEauthorrefmark{2}dat.pham, \IEEEauthorrefmark{4}hoang.nguyen, \IEEEauthorrefmark{5}brain01\}@vinbrain.net \\
\IEEEauthorrefmark{3}kien.dao@aivicam.net
}
\IEEEauthorblockA{\IEEEauthorrefmark{6}These authors contributed equally to this work}
}

\maketitle


\begin{abstract}
Intra-sentential code-switching (CS) refers to the alternation between languages that happens within a single utterance and is a significant challenge for Automatic Speech Recognition (ASR) systems. For example, when a Vietnamese speaker uses foreign proper names or specialized terms within their speech. ASR systems often struggle to accurately transcribe intra-sentential CS due to their training on monolingual data and the unpredictable nature of CS. This issue is even more pronounced for low-resource languages, where limited data availability hinders the development of robust models. In this study, we propose AdaCS, a normalization model integrates an adaptive bias attention module (BAM) into encoder-decoder network. This novel approach provides a robust solution to CS ASR in unseen domains, thereby significantly enhancing our contribution to the field. By utilizing BAM to both identify and normalize CS phrases, AdaCS enhances its adaptive capabilities with a biased list of words provided during inference. Our method demonstrates impressive performance and the ability to handle unseen CS phrases across various domains. Experiments show that AdaCS outperforms previous state-of-the-art method on Vietnamese CS ASR normalization by considerable WER reduction of 56.2\% and 36.8\% on the two proposed test sets.
\end{abstract}


\section{Introduction}

Automatic Speech Recognition (ASR) systems have made great progress in monolingual speech but struggle with intra-sentential code-switching (CS), where speakers alternate between languages within an utterance. This issue is prevalent these days, especially when the speaker mentions foreign-named entities or terminologies in different domains when they do not have a corresponding word in the native language.

The challenge of CS speech recognition primarily stems from the limited data available, which is insufficient to cover all speech variations, especially terms not present in the training data. This issue is even more complex for low-resource languages such as Vietnamese, particularly in the context of Vietnamese medical conversations. These conversations frequently utilize CS due to a majority of international standard medical terms not being translated into Vietnamese. 

It's evident that training an End-to-End (E2E) ASR model, which can transcribe CS utterances directly from acoustic signals to written text, necessitates a comprehensive collection of CS speech data. Prior works have aimed to tackle the issue of intra-sentential CS by incorporating language information to take advantage of the available large scale of text data mainly including two approaches. The first is integrating a language model into the decoding scheme of the E2E ASR system \cite{ct1,ct2,ct3,ct4,ct5}. The second approach \cite{ct6, ct7, ct8, ct9, ct10} proposes the use of correction modules on top of the ASR system to standardize the its output.

Of the methods above, ``plug-in'' modules added to the ASR system demonstrate significant promise in terms of quality and do not require much data or resources for training. Notably, \cite{ct8, ct9, ct11} proposes models that can adapt to new domains during inference by introducing a contextual biasing mechanism through a predefined list of biased words. In this approach, a tagger module is used to identify CS phrases before normalization. However, \cite{ct11} shows inefficiency due to decoding latency, which depends on the number of words in the bias list, and \cite{ct8, ct9} encounters issues with degraded performance with long contextual biasing because of the inadaptability of the tagger module when the bias list changes.


To address this gap, we proposed a novel model called AdaCS, which has the ability to adapt according to a predefined bias list in both CS phrase identification and normalization. This is achieved by utilizing the bias attention module in both phases. To demonstrate the performance and adaptive capability of our proposed model, we constructed a dataset for intra-sentential CS for Vietnamese (a low-resource language) with a total of 50,000 general-domain examples for the development set and 4,000 examples for the evaluation set including general and medical domains.

In summary, our contributions are as follows: (1) A novel model, AdaCS, that can adaptively and effectively address the intra-sentential CS normalization problem. (2) A high quality dataset, including a training and test set, to promote related research and serve as a benchmark for normalizing intra-sentential CS for Vietnamese.


\section{Related Work}


The study of CS in ASR has been a significant focus for scholars. Encoder-decoder attention-based ASR has been transformative in the field, providing impressive results in multilingual ASR systems such as Whisper\cite{ct13}, XLS-R\cite{ct14}, USM\cite{ct15}. However, these systems require significant data for training and their ability to manage CS is not fully clear.



Some researchers have enriched models with language information by training a language identification module \cite{ct16, ct17, ct18, ct19, ct20, ct21}. Others have modified the architecture of ASR \cite{ct2, ct22, ct23} by adding a context encoder that incorporates contextual information into ASR systems. Some have incorporated an external contextual language model \cite{ct1, ct2, ct3, ct4} into the ASR decoding framework to adjust the recognition results toward a context phrase list. However, these methods can slow the system or modify the ASR model's behavior.



An alternative approach is designing normalization modules on top of the ASR system to correct its output. These models, trained with text inputs and outputs, can be obtained on a large scale. Some use a tagger \cite{ct9, ct10, ct11, ct12} to detect CS phrases that need normalization. In \cite{ct9}, the Tagger module functions solely as a classifier, lacking adaptability, making its performance dependent on consistency with the CS bias list. Conversely, \cite{ct24} uses an adaptive tagger only for replacement, relying entirely on the tagger for normalization and skipping it in cases of conflict.

\section{Method}

\begin{figure*}[htbp]
    \centerline{
    \includegraphics[width=1\linewidth]{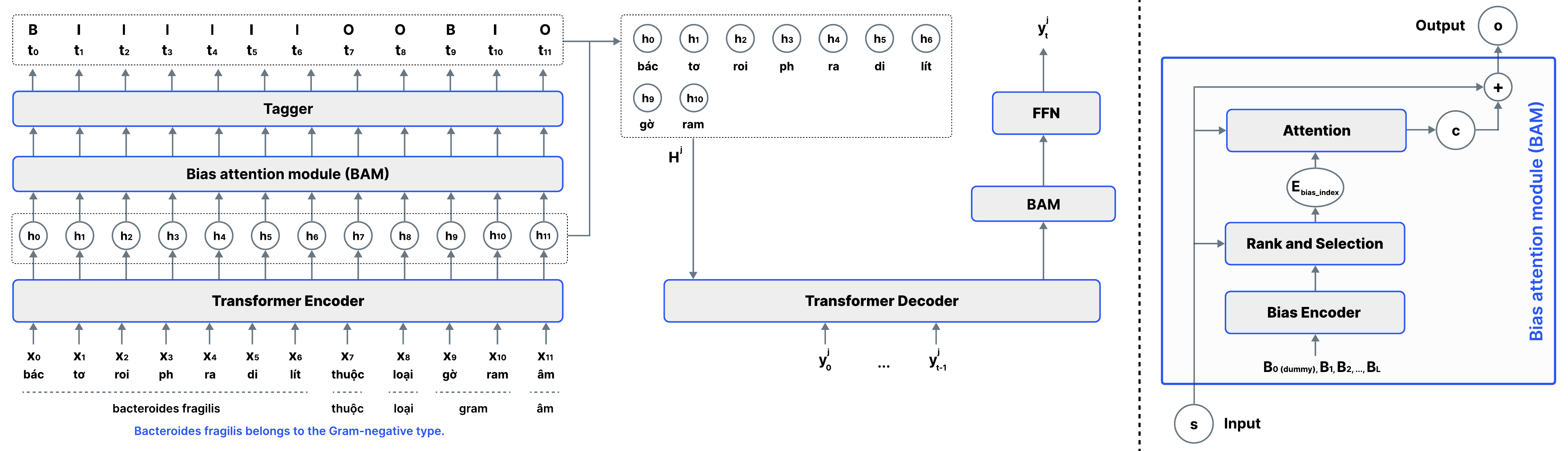}}
    \caption{An overview of the AdaCS architecture, along with an illustrative example. The Bias Attention Module (BAM) is on the right side of the figure, and the Encode-Decode process of AdaCS is on the left side.}
    \label{fig:model_arch}
\end{figure*}

\begin{figure}[htbp]
    \centerline{
    \includegraphics[width=1\linewidth]{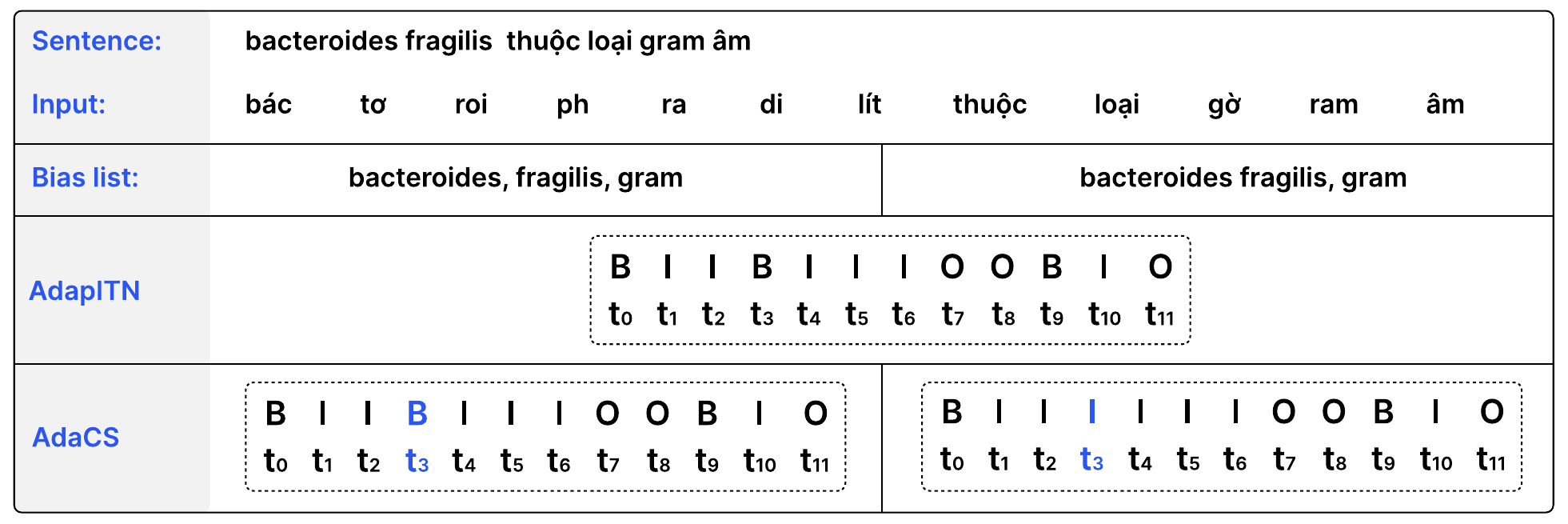}}
    \caption{An example of the impact of the word bias list and phrase bias list on the Tagger of AdaCS and AdapITN. Given the same input sentence, AdapITN produces consistent tagging results whether the bias list being in words (left) or phrases (right). In contrast, AdaCS adapts its tagging accordingly.}
    \label{fig:tagger}
\end{figure}

In this section, we propose a novel model \textbf{Ad}aptive \textbf{C}ode \textbf{S}witching (AdaCS). AdaCS comprises a bias attention module, an encoder, and a decoder. Both the encoder and decoder blocks in AdaCS are integrated with the bias attention module to aid in the accurate and efficient identification and normalization of CS phrases, respectively. An overview of our proposed model is illustrated in Fig. \ref{fig:model_arch}.

\subsection{Bias Attention Module}



The Bias Attention Module (BAM) takes in the hidden representation $s \in \mathbb{R}^{d_{model}}$ of a token and enhances its information about the bias list. BAM consists of: Bias Encoder, Rank \& Selection, and Attention submodules.


The Bias Encoder processes a predefined list of biases, denoted as \(\{B_i\}_{i=1}^L\), where \(L\) is the total number of entries. Each \(B_i\) represents either a single word or a phrase, with \(b_i\) denoting the number of tokens. A dummy entry \(B_0\) is added to handle cases without bias information. For each \(B_i\), the encoder computes a token-level representation matrix 
$
E_i = [e_1, e_2, \ldots, e_{b_i}] \in \mathbb{R}^{b_i \times d_{\text{model}}},
$
and a pooled vector representation 
$
P_i \in \mathbb{R}^{d_{\text{model}}}.
$
Given that both the bias list of CS phrases and the ASR hypothesis are text-based, it is logical to share parameters between the bias encoder and text encoder.

The similarity score of the tokens and the bias phrases is calculated by an inner product operation:

\begin{equation}
    score = sP^{T}
\end{equation}
where $P \in \mathbb{R}^{(L+1)*d_{model}}$ is the bias pooling matrices.

The bias phrase corresponding to the token, which is used to add the information to it, is retrieved by:
\begin{equation}
bias\_index = argmax(scores)
\end{equation}

Next, the bias mechanism is applied by using an attention layer with the query being $s$ and the keys and values being $E_{bias\_index}$ to compute a combined feature $c$. This feature is then summed with the original representation $s$ to form an information-augmented output $o$:

\begin{equation}
    o = s + MultiHeadAttention(s, E_{bias\_index})
\end{equation}

\subsection{Encoder}


Given a input sequence $x_0, ..., x_n$, the encoder compute the contextual features $H = [h_{0}, ..., h_{n}] \in \mathbb{R}^{n*d_{model}}$. Before being fed into the classification layer to determine the corresponding tag $\hat{\tau}_i$ for each token in the input sequence, the features $H$ is information-augmented with the bias list information through our BAM. The tag labels for each input token consists of $B, I, O$ where $B$ indicates that the token is the start of a tagged region, $I$ if the token stands within a tagged region and $O$ otherwise.
\begin{align}
h'_{i} & = BAM(h_{i}) \\
\hat{\tau}_i & = argmax(W_{tagger}h'_{i}), \quad W_{tagger} \in \mathbb{R}^{3*d_{model}}
\end{align}

\subsection{Decoder}


We adopted a decoder that mirrors the successful implementation proposed in \cite{ct9}. After the tagger module identifies the $m$ text regions requiring normalization, the corresponding region embeddings $H^{j}, j \in [0, m)$ are fed into the decoder to generate the corresponding normalized phrase $Y^{j}$. The decoder take the previous output token $y_{0}^j,...,y_{t-1}^{j}$ and does the cross attention with $H^{j}$ to produce a temporary output feature $o^{j}_{t} \in \mathbb{R}^{d_{model}}$. 
Here, BAM is also applied to help this temporary output feature to convey information about the bias list and the information-augmented output is used for decoder prediction:
\begin{align}
z_{t}^{j} & = BAM(o_{t}^{j}) \\
\hat{y}_{t}^{j} & = W_{ffn}(z_{t}^{j}), \quad W_{ffn} \in \mathbb{R}^{V*d_{model}}
\end{align}
\subsection{Losses}
The loss function $L$ used for training is a sum of four components:
The tagger $L_{tagger}$, encoder biasing ranking $L_{enc\_rank}$, decoder biasing ranking $L_{dec\_rank}$ and the classifier of next predicted token $L_{gen}$.
\begin{align*}
    L_{tagger} &= \frac{1}{n} \sum_{i=0}^{n}CE(\hat{\tau}_i, \tau_i) \\
    L_{enc\_rank} &= \frac{1}{n} \sum_{i=0}^{n}CE(score_{i}, label\_enc\_rank_{i}) \\
    L_{dec\_rank} &= \frac{1}{m*T}\sum_{j=0}^{m}\sum_{t=0}^{T}CE(score_{t}^{j}, label\_dec\_rank_{t}^{j}) \\
    L_{gen} &= \frac{1}{m*T} \sum_{j=0}^{m}\sum_{t=0}^{T}CE(\hat{y}_{t}^{j}, y_{t}^{j}) \\
    L &= \alpha L_{tagger} + \beta L_{enc\_rank} + \gamma L_{dec\_rank} + \delta L_{gen}
\end{align*}
where $T$ is the number of output tokens generated by the decoder. 
$label\_enc\_rank_{i}$ and $label\_dec\_rank_{t}^{j}$ is the index of the bias word corresponding to the $i$-th token and $j$-th tagged region, respectively. In our experiments, we use $\alpha = \beta = \gamma = \delta = 1$.

\section{Experiment}
\subsection{Dataset Preparation}\label{sec:dataset-preparation}

\begin{table}[htbp]
\caption{Examples of reference, input, words and phrases bias list in the general-test and medical-test, at "easy" and "hard" levels.}
\begin{center}
\begin{tabular}{|c|}
\hline
\textbf{test-general - easy sample} \\
\hline
\parbox[t]{8cm}{\textbf{Reference:} Ông học tập thiết kế động cơ cơ bản từ \textit{Chevrolet} và nghiên cứu khung gầm xe tải của \textit{Ford}.} \\
\parbox[t]{8cm}{\textbf{Input:} Ông học tập thiết kế động cơ cơ bản từ \textit{che vô lét} và nghiên cứu khung gầm xe tải của \textit{pho}.} \\
\parbox[t]{8cm}{\textbf{Words bias:} Chevrolet, Ford} \\

\hline
\textbf{test-medical - hard sample} \\
\hline
\parbox[t]{8cm}{\textbf{Reference:} \textit{Botulism Antitoxin Heptavalent} là thuốc giải duy nhất cho những trường hợp nhiễm vi khuẩn \textit{Clostridium botulinum}} \\
\parbox[t]{8cm}{\textbf{Input:} \textit{Bô tu lim an ti tô xin hép ta va len} là thuốc giải duy nhất cho những trường hợp nhiễm vi khuẩn \textit{cờ lo tờ ri đi um bô tu li num}} \\
\parbox[t]{8cm}{\textbf{Phrases bias:} Botulism Antitoxin Heptavalent, Clostridium botulinum} \\
\hline
\end{tabular}
\label{tab:examples}
\end{center}
\end{table}

\begin{table*}[h!tbp]
\caption{Evaluation results based on WER of the models on the test-general and test-medical sets that we proposed.}
\begin{center}
\begin{tabular}{|c|c|c|c|c|c|c|c|c|}
\hline
\multirow{2}{*}{\textbf{Model}} & \multirow{2}{*}{\textbf{Bias type}} & \multicolumn{3}{|c|}{\textbf{test-general}} & \multicolumn{3}{|c|}{\textbf{test-medical}} & \multirow{2}{*}{\textbf{Speed (examples/s) ($\uparrow$)}} \\
\cline{3-8}
 & & \textbf{\textit{N-WER} ($\downarrow$)} & \textbf{\textit{CS-WER} ($\downarrow$)} & \textbf{\textit{WER} ($\downarrow$)} & \textbf{\textit{N-WER} ($\downarrow$)} & \textbf{\textit{CS-WER} ($\downarrow$)} & \textbf{\textit{WER} ($\downarrow$)} & \\
\hline
\multirow{1}{*}{Transformers} 
& None & 15.9 & 73.9 & 28.5 & 29.5 & 86.3 & 37.1 & 6.60 \\
\hline
\multirow{3}{*}{GPT-4o} 
& None & 8.2 & 76.3 & 15.4 & 9.1 & 72.4 & 15.0 & 0.11 \\
\cline{2-9}
& Words & 8.7 & 70.3 & 14.7 & 9.8 & 68.9 & 14.7 & 0.08 \\
\cline{2-9}
& Phrases & 9.0 & 67.9 & 14.8 & 9.8 & 69.0 & 14.7 & 0.08 \\
\hline

\multirow{3}{*}{AdapITN} 
& None & 19.2 & 61.3 & 19.1 & 25.3 & 62.3 & 24.4 & \textbf{25.00} \\
\cline{2-9}
& Words & 3.0 & 33.7 & 6.4 & 3.1 & 43.0 & 7.6 & 14.90 \\
\cline{2-9}
& Phrases & 2.62 & 42.1 & 7.3 & 3.3 & 55.4 & 8.9 & 9.45 \\
\hline

\multirow{3}{*}{AdaCS (ours)} 
& None & 20.7 & 62.5 & 20.2 & 28.1 & 69.3 & 26.4 & 23.60 \\
\cline{2-9}
& Words & 1.4 & 18.6 & 3.3 & \textbf{2.2} & \textbf{29.0} & \textbf{4.8} & 14.70 \\
\cline{2-9}
& Phrases & \textbf{1.2} & \textbf{16.1} & \textbf{2.8} & 3.1 & 50.1 & 7.8 & 8.92 \\
\hline
\multicolumn{9}{l}{N-WER: refers the WER on the words that do not require normalization.} \\
\multicolumn{9}{l}{CS-WER: refers to the WER on the CS normalization.} \\
\multicolumn{9}{l}{WER: refers the error throughout the entire process of normalizing the spoken text output of ASR.} \\
\end{tabular}
\label{tab:result}
\end{center}
\end{table*}


First, we collected text data from diverse domains, segmented it into sentences, and preprocessed it to construct the corpus of original sentences. We then identified and selected sentences with CS phrases. Subsequently, we manually labeled the Vietnamese pronunciation of these CS phrases and replaced them in the original sentences, resulting in the spoken-reference pair (Table \ref{tab:examples}).

For the training process, we filtered and selected a total of 50,000 general-domain spoken-reference pairs. With a total count of approximately 1M text tokens, of which CS phrases constitute 7.5\%. For evaluation, we designed the test sets according to the following criteria: (1) The test sets include two distinct domains, namely test-general for the general domain and test-medical for the medical domain. (2) Both test-general and test-medical have at least 90\% of CS phrases that are not presented in the training set. (3) Test sets include "easy" examples where CS phrases are mixed with Vietnamese words and "hard" examples where CS phrases occur consecutively, common when listing proper names or medications (Table \ref{tab:examples}). These criteria allow us to comprehensively evaluate the model's adaptation ability in both in-domain and cross-domain scenarios compared to the training set. After careful curation, we created test sets comprising 2,000 general-domain and 2,000 medical-domain spoken-reference pairs.

\subsection{Experiment setup}

To evaluate our proposed model, we compared AdaCS with traditional Transformer model\cite{ct25} as the baseline and other models including GPT-4o\cite{ct26} and AdapITN\cite{ct9}.

We propose the following experimental settings. In the first experiment, no bias list is used to test the native normalization ability of the models. The second experiment uses a random bias list with a predetermined size of 1000 CS words, drawn from the list of English words in the entire corresponding test set, combining English words from current sentences to perform normalization. In the third experiment, we aim to reflect real-world conversations, where English phrases consisting of words often used to indicate a concept. We follow a similar approach to the second experiment, but our bias list includes both phrases and words instead of just words. In the final experiment, we evaluate the models' performance as the size of the bias list increases, intending to assess efficiency in a production environment. 


\subsection{Training}


We employ the training dataset as described in \ref{sec:dataset-preparation} to train the baseline model (an encoder-decoder Transformers \cite{ct25}), AdapITN\cite{ct9} and AdaCS. For the mentioned models, we use EnViBERT\cite{ct27,ct28} as the base pretrained model.
For each training step, a bias list is generated that includes the English words present in the sentences within the batch, as well as random English words from the database, so that the total number of bias words is approximately 1000 CS phrases.

\subsection{Result \& Analysis}
\begin{figure}[htbp]
    \centerline{
    \includegraphics[width=1\linewidth]{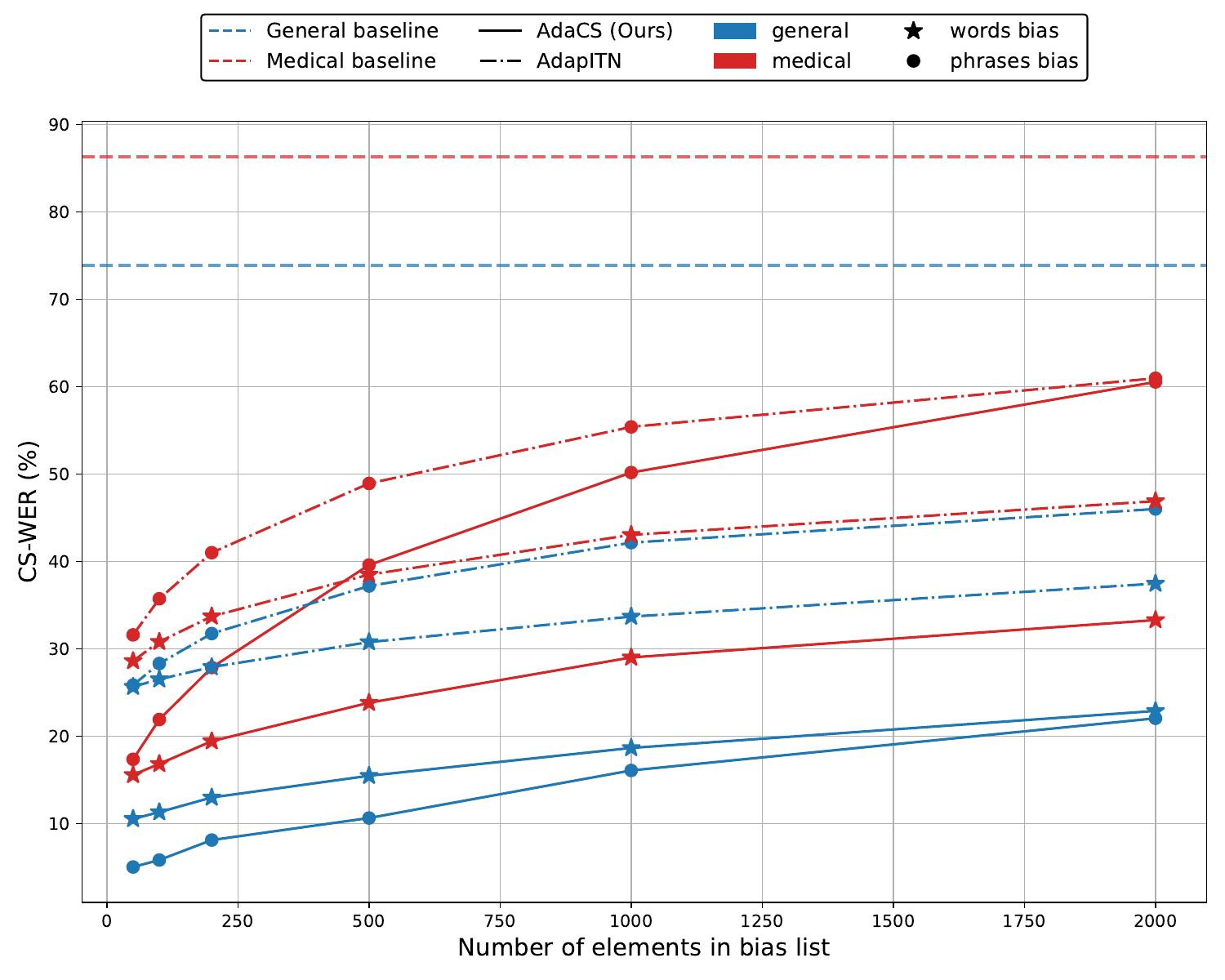}}
    \caption{Performance of AdaCS and AdapITN as the size of the bias list increases on test sets.}
    \label{fig:ex4}
\end{figure}

Table \ref{tab:result} presents the results of the first three experiments. Without a bias list, GPT-4o achieves the best performance, reducing WER by 45.9\% and 59.5\% on the test-general and test-medical datasets, respectively, compared to baseline. AdaCS and AdapITN also show notable improvements, with WER reductions of 29.1\% and 32.9\% on the test-general dataset, and 28.8\% and 34.2\% on the test-medical dataset, respectively, compared to the baseline. Interestingly, using word/phrase biases leads to a significant improvement for AdaCS and AdapITN, outperforming GPT-4o, with relative WER reductions of 46.9\% to 80.9\% compared to the baseline.


AdaCS outperforms AdapITN with word-level bias across all WER metrics on both test sets, showcasing its adaptability with unseen data. Regarding phrase-level bias, there are differing trends in AdaCS between the two test sets. While the general domain exhibits an improvement with a 2.5\% absolute decrease in CS-WER, the medical test shows an increase in WER compared to word-level bias. This can be explained by the fact that phrases in the medical domain often contain similar parts, which creates challenges for both models. Nevertheless, AdaCS continues to outperform AdapITN in these tests in overall with relative improvement of CS-WER from 9.5\% to 61.8\% across the experiment settings when ultilizing bias mechanism, without significant trade-offs in terms of speed. Figure \ref{fig:tagger} is the explanation for this result, where the dynamic tagging by AdaCS when the bias list changes leads to more accurate normalization. 

The correlation between the size of the bias list and the CS-WER metric is demonstrated in Figure \ref{fig:ex4}. In general,the effectiveness of both models that employ context-aware biasing tends to decline as the quantity of bias words rises. However, their CS-WER remains superior to the baseline. Furthermore, it is shown that AdaCS consistently outperforms AdapITN as the number of elements in the bias list changes across both test sets. These experimental results underscore the performance of AdaCS, particularly in handling code-switching and its adaptability within both general and domain-specific datasets.


\section{Conclusion}
AdaCS model demonstrates significant advancements in normalization intra-sentential CS. We believe our proposal emphasizes the importance of leveraging contextual information and tailored biasing strategies to improve CS speech recognition performance. The dataset, checkpoint, and experimental results are available at: https://github.com/adacs-project/adacs-project.github.io/.

\end{document}